\PassOptionsToPackage{table}{xcolor}
\documentclass[]{lightspeed}

\usepackage[utf8]{inputenc}
\usepackage{latexsym}
\usepackage{fvextra}
\usepackage{tabularx}
\usepackage{array}
\usepackage{amssymb}
\usepackage{pifont}
\usepackage{wrapfig}
\usepackage{amsmath,amssymb}
\usepackage{booktabs}
\usepackage{multirow}
\usepackage{graphicx}
\usepackage{tabularx}
\usepackage{makecell}
\usepackage{placeins}
\usepackage{url}
\usepackage{hyperref}

\microtypesetup{expansion=false}
\setcitestyle{square,comma,numbers,sort&compress}

\definecolor{lightgray}{gray}{0.92}

\title{Ex-Omni: Enabling 3D Facial Animation Generation for Omni-modal Large Language Models}

\author[1,*]{Haoyu Zhang}
\author[2]{Zhipeng Li}
\author[3,\dagger]{Yiwen Guo}
\author[1,\dagger]{Tianshu Yu}

\affiliation[1]{The Chinese University of Hong Kong, Shenzhen}
\affiliation[2]{LIGHTSPEED}
\affiliation[3]{Independent Researcher}

\contribution[*]{Work done during an internship at LIGHTSPEED}
\contribution[\dagger]{Corresponding authors}

\email{haoyuzhang3@link.cuhk.edu.cn}
\email{zhipengxli@tencent.com}
\email{guoyiwen89@gmail.com}
\email{yutianshu@cuhk.edu.cn}

\abstract{Omni-modal large language models (OLLMs) aim to unify multimodal understanding and generation, yet extending them to jointly produce speech and 3D facial animation remains largely underexplored. A key challenge is the mismatch between the discrete semantic reasoning of LLMs and the dense temporal dynamics required for 3D facial motion. We propose Expressive Omni (Ex-Omni), a framework that augments OLLMs with speech-accompanied 3D facial animation. Ex-Omni decouples semantic reasoning from temporal generation through a speech-unit generator with blendshape co-supervision and a non-autoregressive blendshape decoder, where speech units provide temporal scaffolding and hidden speech representations carry facially relevant cues. We further introduce a token-as-query gated fusion (TQGF) interface for controlled semantic injection, as well as InstructS2SF-1200K, a 1.2M-sample weakly supervised dataset for speech-accompanied facial animation. Extensive experiments show that Ex-Omni retains competitive speech QA capability while natively generating coordinated text, speech, and 3D facial animation, and approaches the Audio2Face-3D teacher cascade in synchronization and human preference.}

\headercontent{
  \begin{tabular}{c}
    \url{https://github.com/Tencent/Ex-Omni}
  \end{tabular}
}

\date{September 2026}

\begin{document}
\thispagestyle{firstheader}
\maketitle

\section{Introduction}
Large language models (LLMs) have achieved strong progress across a wide range of tasks \cite{minaee2024large}, demonstrating impressive generalization and reasoning capabilities. As research continues to expand from unimodal understanding toward multimodal understanding and generation, growing attention has been devoted to unifying these tasks within a single framework, commonly referred to as omni-modal large language models (OLLMs). With the continued development of such omni models \cite{DBLP:journals/corr/abs-2501-01957, DBLP:journals/corr/abs-2410-11190, DBLP:journals/corr/abs-2511-00279, li2025uni, DBLP:journals/corr/abs-2408-16725, luo2025openomni, DBLP:journals/corr/abs-2503-20215, DBLP:journals/corr/abs-2506-09344}, these systems are increasingly relevant to human-computer interaction and embodied intelligence.

In the area of human-computer interaction, there is a growing demand for OLLMs that can engage in natural and expressive interactions with humans. Human communication is inherently multimodal and further extends beyond verbal content alone. In face-to-face interaction, temporally coherent 3D facial animation synchronized with speech plays a crucial role in conveying non-verbal cues and enhancing interaction naturalness, particularly in applications such as virtual characters, digital avatars, and embodied agents. However, existing OLLMs primarily focus on linguistic, acoustic, or pixel-level visual outputs, while expressive non-verbal modalities such as 3D facial animation remain largely underexplored. Motivated by this gap, we investigate integrating 3D facial animation generation into OLLMs. A natural idea is to directly attach a blendshape decoder to an LLM and predict animation from its hidden representations. In practice, we found this design exposes a challenge: LLM hidden states are optimized for sparse, token-level semantics with weakly constrained temporal structure, whereas 3D facial animation requires dense and temporally smooth motion at a much finer time scale. Bridging these representations forces the decoder to infer fine-grained dynamics from coarse semantic features, resulting in an ill-conditioned mapping that typically demands substantially larger model capacity and more paired speech–face supervision for stable generation.

In this paper, we propose Expressive Omni (Ex-Omni), an omni-modal framework that augments OLLMs with speech-accompanied 3D facial animation, where facial motion is represented using ARKit-52 blendshape coefficients~\cite{DBLP:conf/eurographics/LewisARZPD14} and generated in a non-autoregressive manner. Ex-Omni follows text/speech instructions to generate synchronized speech paired with facial animation in an end-to-end manner. We use \emph{expressive} in the sense common to speech-driven facial animation: coherent articulation, phoneme-related mouth motion, and speech--face coordination in addition to the spoken content. To facilitate stable and temporally coherent facial animation learning from LLM semantics, Ex-Omni decouples semantic reasoning from temporal generation through a two-stage design. Rather than directly predicting facial motion from LLM hidden states, Ex-Omni first employs a speech-unit generator with blendshape co-supervision, where discrete speech units provide explicit temporal scaffolding while the generator hidden states are encouraged to encode facially relevant cues. A blendshape decoder then predicts blendshape sequences conditioned on both signals. To better bridge high-level semantics and temporal motion generation, we further introduce a unified token-as-query gated fusion (TQGF) mechanism to selectively regulate how and when semantic information is injected into the speech and facial generation processes, simplifying optimization and improving temporal alignment.

We also construct InstructS2SF-1200K, a 1.2M-sample weakly supervised dataset for training OLLMs with speech-accompanied 3D facial animation. It contains two subsets tailored to different training stages: 1M Text-to-Speech-Face (TTSF) samples for speech-blendshape co-pretraining, and 200K Speech-to-Speech-Face (S2SF) QA samples for dialogue-based speech-blendshape co-pretraining. By combining large-scale weak supervision with speech-face data, InstructS2SF-1200K helps bridge the gap between limited real-world recordings and open-domain generalization. Experimental results show that Ex-Omni preserves substantial speech-task capability, improves over the evaluated non-teacher cascades in the controlled downstream setting, and remains close to the Audio2Face-3D teacher cascade while supporting native face generation.

Overall, the key contributions are:
\begin{itemize}
  \item We propose Expressive Omni (Ex-Omni), an OLLM framework for unified instruction following and generation across text, speech, and speech-accompanied 3D facial animation.

  \item To reduce the difficulty of learning temporally coherent 3D facial animation from LLM semantics, Ex-Omni adopts a two-stage temporal-generation design. Discrete speech units provide the primary temporal scaffold, while a token-as-query gated fusion (TQGF) module injects contextual speech representations into the blendshape decoder.

  \item We construct InstructS2SF-1200K, a 1.2M-sample weakly supervised corpus with 1M TTSF synthesis samples and 200K dialogue-based S2SF samples. Experiments show that Ex-Omni approaches its teacher cascade and improves over the evaluated non-teacher cascades in controlled downstream comparisons while reducing face-branch latency.

\end{itemize}

\section{Related Work}
\paragraph{Omni-modal Large Language Models.} OLLMs \cite{DBLP:journals/corr/abs-2501-01957, DBLP:journals/corr/abs-2410-11190, DBLP:journals/corr/abs-2511-00279, li2025uni} represent an important direction in multimodal large language models, as they integrate the capabilities of understanding and generating across multiple modalities, such as text, speech, and vision, within a unified framework. For example, Mini-Omni \cite{DBLP:journals/corr/abs-2408-16725} utilizes text-instructed speech generation and batch-parallel strategies to achieve seamless speech synthesis while preserving the model’s text capabilities. OpenOmni \cite{luo2025openomni} introduces a two-stage training framework to achieve zero-shot cross-modal alignment from vision-language tasks to speech-language tasks. Qwen2.5-Omni \cite{DBLP:journals/corr/abs-2503-20215} introduce the Thinker-Talker architecture to integrate text, speech, and vision modalities into a unified end-to-end model. Ming-Omni \cite{DBLP:journals/corr/abs-2506-09344} integrates visual generation capabilities into a unified omni-modal model, utilizing modality-specific routers to achieve understanding and generation across multiple modalities.

\paragraph{Facial Animation Generation.} Facial animation generation is an important research area for improving system interactivity. Earlier methods \cite{DBLP:conf/cvpr/ChenMDX19, DBLP:conf/wacv/MittalW20, DBLP:conf/iccv/ZhangZHZNBG21, DBLP:conf/cvpr/HongZS022} mainly focus on 2D facial animation generation, a field that has become mature after years of research. In recent years, 3D facial animation generation \cite{DBLP:conf/iccv/RichardZWTS21, DBLP:conf/cvpr/XingXZC0W23, DBLP:conf/iccv/PengWSXZH0F23, DBLP:conf/mm/PengLSXZLHF23, DBLP:conf/eccv/FanLLXY24, DBLP:conf/cvpr/PengFWW0HF25} has gradually received more attention. These methods have generally focused on predictions based on mesh representations or parameterized models to enhance realism. For mesh-based methods, FaceFormer \cite{DBLP:conf/cvpr/FanLSWK22} introduces a transformer-based approach for generating 3D facial animations, using autoregressive modeling to capture long-term audio context. CodeTalker \cite{DBLP:conf/cvpr/XingXZC0W23} utilizes discrete motion priors learned from real facial movements, applying a vector quantized autoencoder to reduce uncertainty in the audio-to-motion mapping process. For parameterized methods, ARKit-like blendshape models \cite{DBLP:conf/eurographics/LewisARZPD14} are commonly used. EmoTalk \cite{DBLP:conf/iccv/PengWSXZH0F23} disentangles emotion and content from speech to generate expressive facial movements, while DuelTalk \cite{DBLP:conf/cvpr/PengFWW0HF25} supports multi-round, dual-speaker interactions. UniTalker \cite{DBLP:conf/eccv/FanLLXY24} combines mesh-based and parameterized annotation styles to improve scalability, and xADA \cite{taylor2025xada} generates controllable face, tongue, and head animation directly from input speech. These systems study the standalone audio-to-animation mapping. Ex-Omni instead studies how facial animation can be generated natively alongside a newly reasoned text and speech response inside an OLLM.

\paragraph{Speech Language Models.} Recent advancements in speech language models \cite{DBLP:conf/iclr/FangGZMZ025, DBLP:conf/acl/FangZGZ025, DBLP:journals/corr/abs-2501-06282, DBLP:conf/emnlp/ZhangLZZWZQ23, DBLP:conf/nips/HassidRNGCKCDSD23, DBLP:journals/corr/abs-2407-10759, DBLP:conf/acl/Chen0YLLXN00L0025, DBLP:journals/corr/abs-2408-16725} have significantly improved speech understanding and generation in an end-to-end manner, eliminating the need for cascaded ASR and TTS models. For example, SpeechT5 \cite{DBLP:conf/acl/AoWZ0RW0KLZWQ0W22} aligns text and speech representations into a shared semantic space using a unified encoder-decoder structure and cross-modal vector quantization methods. Moshi \cite{DBLP:journals/corr/abs-2410-00037} addresses the issues of latency and information bottlenecks in traditional speech dialogue systems through its full-duplex speech-to-speech generation framework and the Inner Monologue design. SpeechGPT-Gen \cite{DBLP:journals/corr/abs-2401-13527} introduces the Chain-of-Information Generation to decouples the modeling of semantic and perceptual information,thus making the speech generation process more efficient and precise. GLM-4-Voice \cite{DBLP:journals/corr/abs-2412-02612} addresses the delay and error accumulation problems by adopting a 12.5Hz speech segmenter, streaming reasoning and large-scale speech-to-text pre-training.

\section{Method}
\subsection{Overview}
\begin{figure}[htbp]
 \centering
 \includegraphics[width=0.60\linewidth]{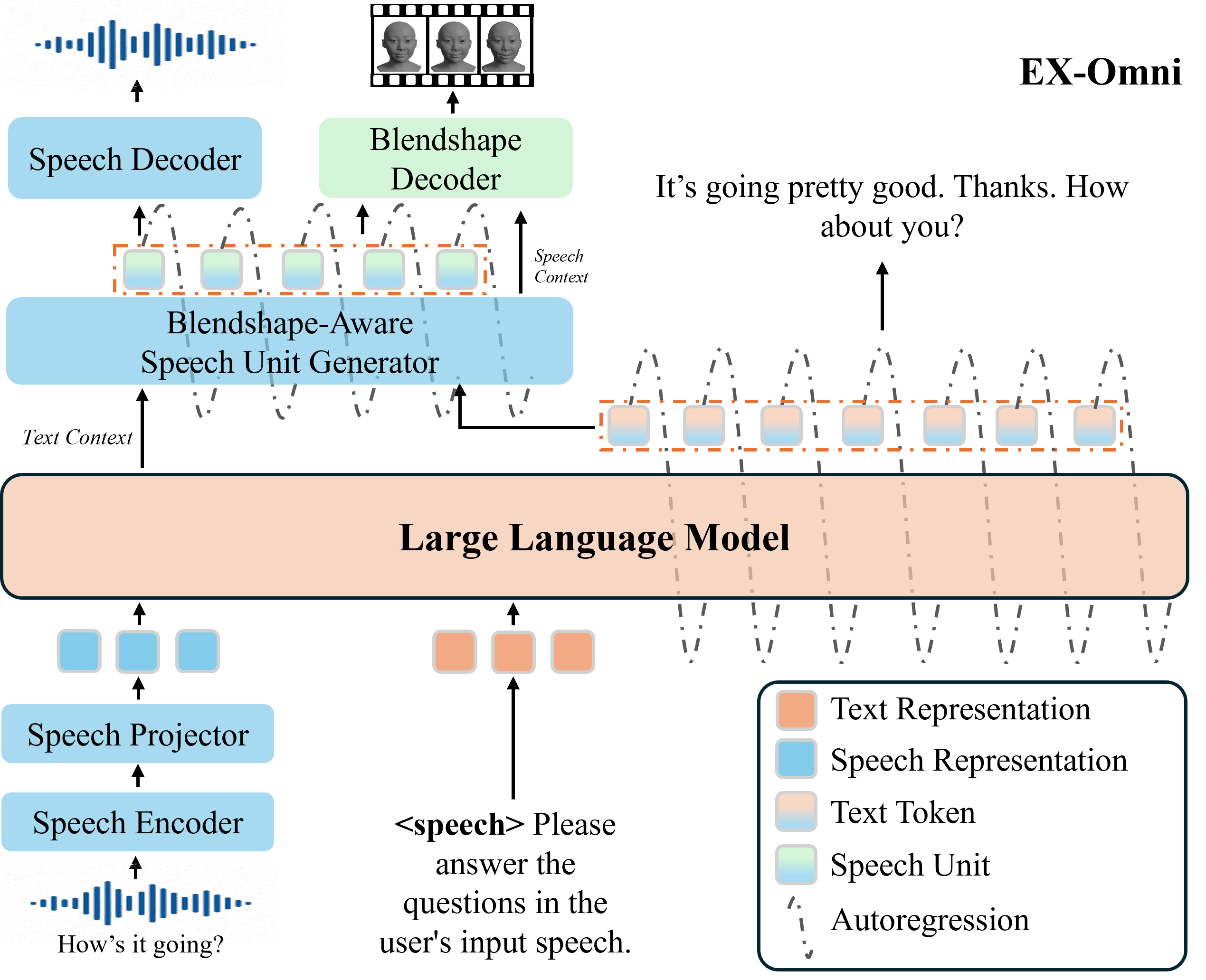}
 \caption{Model architecture of Ex-Omni.}
 \label{fig:pipeline}
 \end{figure}

Figure~\ref{fig:pipeline} shows the overall pipeline of Ex-Omni. Given an instruction $\mathcal{I}$ comprising a text input $x$, a speech waveform $a$, or both, Ex-Omni performs instruction understanding and multimodal generation within an LLM-centered unified framework, where the LLM focuses on semantic reasoning rather than direct temporal generation.
The model produces a discrete speech unit sequence $u$, which is decoded into waveform speech, as well as a sequence of 3D facial animation parameters $y$ (\textit{i.e.,} blendshape coefficients). At a high level, Ex-Omni adopts a structured decomposition that reduces the difficulty of learning temporally coherent generation by decoupling semantic reasoning from modality-specific temporal synthesis. Specifically, the LLM is responsible for instruction understanding and semantic reasoning, while speech units are used as an explicit temporal scaffolding to guide downstream speech and facial animation generation. We formulate the overall model as
\begin{equation}
(u, y) = \mathcal{F}(\mathcal{I}; \theta), \quad
\theta = \{ \theta_E, \theta_P, \theta_L, \theta_U, \theta_F \},
\end{equation}
where $\theta_E$, $\theta_P$, and $\theta_L$ correspond to the speech encoder, speech projector and LLM, respectively.
$\theta_U$ denotes the speech-unit generator with blendshape co-supervision while $\theta_F$ represents the blendshape decoder.

\subsection{Unified Speech-Text Representation}
Given an instruction $\mathcal{I}$ containing a text input $x$, a speech waveform $a$, or both, we map each available input modality into the shared LLM token embedding space.
When speech is present, it is first encoded into high-level representations and then projected as
\begin{equation}
X_s = P_{\theta_P}(E_{\theta_E}(a)) \in \mathbb{R}^{T_s \times d},
\end{equation}
while the available text tokens are embedded as
\begin{equation}
X_l = \mathrm{Emb}_{\theta_L}(x) \in \mathbb{R}^{T_l \times d},
\end{equation}
where $d$ denotes the LLM embedding dimension and $\mathrm{Emb}_{\theta_L}$ is the input token embedding layer of Qwen3-8B \citep{DBLP:journals/corr/abs-2505-09388}.
Depending on the available input modalities, the unified input representation is constructed as
\begin{equation}
X =
\begin{cases}
X_l, & \text{if $\mathcal{I}$ contains text only},\\
X_s, & \text{if $\mathcal{I}$ contains speech only},\\
[X_l; X_s], & \text{if $\mathcal{I}$ contains both modalities},
\end{cases}
\quad
X \in \mathbb{R}^{T_X \times d},
\end{equation}
where $T_X$ denotes the resulting input sequence length.

\subsection{LLM-Centered Reasoning}
The LLM serves as a semantic reasoner that focuses exclusively on instruction understanding and high-level reasoning. Given the unified input representation $X$, the LLM (Qwen3-8B) performs autoregressive generation to produce the text response $t_{1:T_{lr}}$.
During this process, we extract the sequence of last hidden states corresponding to the generated response tokens:
\begin{equation}
H = [h_1, h_2, \dots, h_{T_{lr}}] \in \mathbb{R}^{T_{lr} \times d},
\end{equation}
where $h_i$ represents the features at step $i$ containing high-level semantic reasoning information. The probability of the next token is predicted based on these states:
\begin{equation}
p_{\theta_L}(t_{i+1} \mid t_{1:i}, X) = \mathrm{Softmax}(W_o h_i).
\end{equation}

\subsection{Joint Speech and 3D Facial Animation Generation}
\label{sec:sp_face_generator}
Ex-Omni jointly generates speech and 3D facial animation, aiming to maintain semantic consistency and temporal alignment across modalities. We adopt token-as-query gated fusion (TQGF), a task-specific cross-attention module in which the target token sequence serves as the query and upstream semantic representations provide contextual key/value states. This design assigns temporal responsibility to the target token sequence and injects semantic cues through a learned gate, preserving the speech-unit temporal scaffold while avoiding direct dense-motion regression from LLM states.

Formally, let $Q \in \mathbb{R}^{M \times d}$ denote query tokens and $C \in \mathbb{R}^{N \times d_c}$ denote context tokens. We first project $C$ to the module width $d$ when needed. The gated fusion operation is
\begin{equation}
\mathrm{Fuse}(Q, C) = Q + \sigma(QW_g+b_g)\odot \mathrm{Attn}(Q, P_C(C)),
\end{equation}
where $\mathrm{Attn}(Q,P_C(C))$ is cross-attention from $Q$ to the projected context, $W_g$ and $b_g$ produce element-wise gates with the same hidden width as the attention output, and $\sigma(\cdot)$ is the sigmoid function. In multi-head attention the gate is applied per token and broadcast over the corresponding head dimensions.

For speech generation, we model speech synthesis as an autoregressive prediction of discrete speech units using a speech-unit generator with blendshape co-supervision (Qwen3-0.6B). Given the generated text tokens $t_{1:T_{lr}}$, the speech-unit generator with blendshape co-supervision enriches the explicit token embeddings with the semantic hidden states $H$ through:
\begin{equation}
\tilde{H} = \mathrm{Fuse}_{\theta_U}\left(\mathrm{Emb}_{\theta_U}(t_{1:T_{lr}}), P_U(H)\right),
\end{equation}
where $\mathrm{Emb}_{\theta_U}$ denotes the token embedding layer adopted from GLM-4-Voice, $P_U$ maps LLM hidden states to the unit-generator width, and $\tilde{H} \in \mathbb{R}^{T_{lr} \times d}$ serves as the conditioning signal. Then, the speech-unit generator with blendshape co-supervision predicts a sequence of speech units $u_{1:T_u}$:
\begin{equation}
p_{\theta_U}(u_{1:T_u} \mid \tilde{H})
= \prod_{j=1}^{T_u} p_{\theta_U}(u_j \mid u_{<j}, \tilde{H}).
\end{equation}

For 3D facial animation generation, we parameterize facial motion using ARKit-52 blendshape coefficients and formulate S2F generation as a non-autoregressive sequence prediction problem, where the model outputs the full blendshape coefficients $\hat{y}_{1:T_y} \in \mathbb{R}^{T_y \times 52}$ in parallel.
Given discrete speech units $u_{1:T_u}$, we first obtain unit embeddings and align them to the target video frame rate by temporally resampling features to length $T_y$ (linear interpolation in our implementation), yielding frame-level query representations $Q_y \in \mathbb{R}^{T_y \times d}$. In parallel, we project the hidden states of the speech-unit generator with blendshape co-supervision into the same space to obtain contextual key/value representations $S \in \mathbb{R}^{T_u \times d}$, which carry semantically rich information. We then inject speech semantics into the frame-level queries using the TQGF module:
\begin{equation}
H_{f} = \mathrm{Fuse}_{\theta_F}(Q_y, S) \in \mathbb{R}^{T_y \times d}.
\end{equation}
We then apply rotary positional embeddings \citep{DBLP:journals/ijon/SuALPBL24} to $H_{f}$ and refine it with a non-autoregressive Transformer decoder, thus obtaining the predicted sequence of 3D facial parameters:
\begin{equation}
\hat{y}_{1:T_y} = D_{\theta_F}(H_{f}), \quad \hat{y}_t \in \mathbb{R}^{52}.
\end{equation}

\subsection{Training Objectives}
\label{sec:training}
\paragraph{Autoregressive Objectives for Text and Speech.}
For text tokens and discrete speech units, we adopt standard autoregressive modeling. Given a token sequence $z = (z_1, \ldots, z_T)$, the objective is $\mathcal{L}_{\text{ar}} = - \sum_{t=1}^{T} \log p(z_t \mid z_{<t})$, 
where $z_t$ denotes either a text token or a speech unit, depending on the supervision available for a given sample. In practice, the text-token objective supervises ASR transcripts and response text in S2SF samples, while the speech-unit objective is applied to both TTSF and S2SF samples.

\paragraph{3D Facial Animation Generation.}
For 3D facial animation, the blendshape decoder is trained with a frame-wise regression loss. Let $\hat{y}_{t} \in \mathbb{R}^{K}$ and $y_{t} \in \mathbb{R}^{K}$ denote the predicted and ground-truth blendshape annotations at frame $t$, respectively. We define the facial loss as $\mathcal{L}_{\text{bs}} = \frac{1}{B} \sum_{i=1}^{B} \frac{1}{|\mathcal{T}_i|} \sum_{t \in \mathcal{T}_i} \left\| \hat{y}_{t}^{(i)} - y_{t}^{(i)} \right\|_2^2$, where $\mathcal{T}_i$ denotes the valid temporal range determined by the target sequence length.

\subsection{Training Strategy}
\label{sec:staged_training}
\paragraph{Stage I (Speech-to-Text Pretraining).}
We train the speech projector on ASR data while freezing all other components.
This stage aligns speech representations with the semantic space of the base LLM.

\paragraph{Stage II (Speech-Blendshape Co-pretraining).}
We pre-train the speech-unit generator and the blendshape decoder on 1M TTSF samples while freezing the base LLM and unrelated modules. This stage introduces joint speech and blendshape supervision to establish speech-blendshape alignment; the unit generator becomes blendshape-aware through gradients from the jointly trained decoder rather than through a separate explicit blendshape target.

\paragraph{Stage III (Dialogue-based Speech-Blendshape Co-pretraining).}
The LLM, speech projector, speech-unit generator with blendshape co-supervision, and blendshape decoder are jointly optimized on a mixture of ASR, TTSF, and S2SF data, while the speech encoder and speech decoder remain frozen. Text autoregression, speech-unit autoregression, and blendshape regression are computed in one masked graph and applied only when the corresponding labels are present. Gradients update the blendshape decoder and speech-unit generator with blendshape co-supervision and, through their conditioning states, the unfrozen LLM and speech projector at substantially smaller learning rates. Ex-Omni predicts speech units and blendshapes; the frozen waveform decoder is not trained. This stage adapts the model to dialogue-oriented speech-blendshape generation.

\section{Data Construction}

\begin{wraptable}{r}{0.48\textwidth}
\vspace{-1.1\baselineskip}
\centering
\caption{Statistics of the three-stage training corpus.}
\label{tab:datasets}
\scriptsize
\setlength{\tabcolsep}{3pt}
\resizebox{\linewidth}{!}{
\begin{tabular}{c c r r r r r r}
\toprule
\multirow{2}{*}{Stage}
& \multirow{2}{*}{Type}
& \multicolumn{6}{c}{Duration (Hour / Second)} \\
\cmidrule(lr){3-8}
& & Total & Mean & Median & Min & Max & Std \\
\midrule
I & ASR
& 2113.38 & 10.47 & 10.46 & 0.83 & 30.00 & 4.81 \\
\midrule
II & TTSF
& 2814.57 & 10.13 & 8.68 & 0.40 & 161.80 & 5.71 \\
\midrule
\multirow{4}{*}{III}
& S2SF (Prompt)
& 782.73 & 14.09 & 12.68 & 1.16 & 75.36 & 6.61 \\
& S2SF (Response)
& 1434.67 & 25.82 & 23.04 & 0.40 & 296.40 & 14.31 \\
& ASR
& 25.91 & 9.33 & 8.07 & 3.00 & 30.00 & 5.01 \\
& TTSF
& 28.38 & 10.22 & 8.72 & 0.64 & 52.20 & 5.76 \\
\bottomrule
\end{tabular}
}
\vspace{-0.5\baselineskip}
\end{wraptable}

The data statistics of the full training corpus used by Ex-Omni are summarized in Table~\ref{tab:datasets}. Overall, the training data consist of an external ASR corpus together with our proposed InstructS2SF-1200K. Large-scale paired text, speech, and high-quality 3D facial motion for open-domain OLLM training is difficult to collect, so we use Audio2Face-3D \citep{DBLP:journals/corr/abs-2508-16401} pseudo-labels as weak supervision. This follows the broader use of synthesized targets as a practical scaling strategy in OLLM training~\cite{DBLP:conf/iclr/FangGZMZ025,luo2025openomni}. Because speech-driven facial motion is inherently one-to-many, these pseudo-labels provide a consistent temporal supervision signal rather than a unique facial-motion ground truth. More details can be found in Appendix~\ref{appendix: data_construction}.

\section{Experiments}
\subsection{Evaluation}
\paragraph{Speech-/Text-to-Face Evaluation.}
For S2F, evaluations are conducted on AlpacaEval \cite{li2023alpacaeval} and CommonEval subsets of VoiceBench~\citep{DBLP:journals/corr/abs-2410-17196}. We use Sync-C and Sync-D from SyncNet \cite{Chung16a, 10.1145/3394171.3413532} (higher Sync-C and lower Sync-D indicate better alignment) for lip-speech synchronization. SyncNet is a widely used third-party reference model for audio-visual synchronization evaluation and is independent of the Audio2Face-3D teacher used for pseudo-label generation. Lip Vertex Error (LVE) requires a paired ground-truth facial-motion sequence, which open-domain generated dialogue does not provide; computing LVE against a teacher output would turn the pseudo-label source into the metric target. We therefore combine SyncNet with a controlled downstream comparison, a human preference study, temporal motion diagnostics, and qualitative videos. T2F evaluation follows the same protocol as S2F evaluation, except that the input is text rather than speech. 

\paragraph{Speech QA Evaluation.}
We evaluate speech QA on VoiceBench \cite{DBLP:journals/corr/abs-2410-17196}, which covers a diverse set of speech-based tasks, including open-ended question answering, reference-based QA, multiple-choice QA, reasoning, instruction following and safety. Open-ended QA is evaluated using GPT-based scores (scores from 1-5), while other tasks are evaluated using accuracy-based metrics. All the evaluations were conducted using the open-source code of VoiceBench to ensure consistency. 

\paragraph{Speech Generation Evaluation.}
For speech generation evaluation, we assess both response-level speech quality and speech-text consistency on the AlpacaEval and CommonEval speech QA sets. Specifically, we compute UTMOS on the generated speech, and transcribe the generated waveform using Whisper-Large-V3 \cite{DBLP:conf/icml/RadfordKXBMS23}. The resulting ASR transcript is then compared with the model's textual response using Word Error Rate (WER), which measures consistency between spoken and textual outputs.

\paragraph{Baselines}
For S2F evaluation, we compare Ex-Omni with two recent S2F methods, EmoTalk~\cite{DBLP:conf/iccv/PengWSXZH0F23} and UniTalker~\cite{DBLP:conf/eccv/FanLLXY24}, both of which support direct prediction of facial blendshape coefficients. We additionally include Ex-Omni+Audio2Face-3D as a teacher-cascade baseline because Audio2Face-3D supplies the weak animation supervision used to train Ex-Omni. All Ex-Omni cascades receive exactly the same generated text and speech as Ex-Omni, isolating the facial animation module. For speech QA evaluation, we compare Ex-Omni with several representative OLLMs and speech large language models. Specifically, the baselines include Qwen2.5-Omni \cite{DBLP:journals/corr/abs-2503-20215}, VITA-1.0 \cite{DBLP:journals/corr/abs-2408-05211}, VITA-1.5 \cite{DBLP:journals/corr/abs-2501-01957}, Mini-Omni \cite{DBLP:journals/corr/abs-2408-16725}, Mini-Omni2 \cite{DBLP:journals/corr/abs-2410-11190}, Moshi \cite{DBLP:journals/corr/abs-2410-00037}, SLAM-omni \cite{DBLP:conf/acl/Chen0YLLXN00L0025}, and LLaMA-Omni \cite{DBLP:conf/iclr/FangGZMZ025}. For speech generation evaluation, we compare Ex-Omni with Qwen2.5-Omni under the same response-generation setting, and report both UTMOS \citep{DBLP:conf/interspeech/SaekiXNKTS22} and speech-text consistency measured by WER.

\begin{table}[htbp]
\centering
\scriptsize
\setlength{\tabcolsep}{8pt}

\caption{End-to-end comparison with non-teacher cascaded OLLM+S2F systems.
Cascaded methods pass generated speech to an external S2F model, whereas
Ex-Omni jointly generates speech and facial animation;
Table~\ref{tab:controlled_downstream} isolates the downstream module under
fixed speech. The best result in each column is shown in \textbf{bold}.}
\label{tab:speech2face}

\begin{tabularx}{\textwidth}{
    >{\raggedright\arraybackslash}X
    c c c c
}
\toprule
\multirow{2}{*}{Method}
& \multicolumn{2}{c}{
    Speech-to-Face
    (Sync-D $\downarrow$ / Sync-C $\uparrow$)}
& \multicolumn{2}{c}{
    Text-to-Face
    (Sync-D $\downarrow$ / Sync-C $\uparrow$)}
\\
\cmidrule(lr){2-3}
\cmidrule(lr){4-5}
& AlpacaEval
& CommonEval
& AlpacaEval
& CommonEval
\\
\midrule

Qwen2.5-Omni-3B + EmoTalk
& 10.538 / 3.305
& 10.750 / 3.203
& 10.643 / 3.166
& 10.830 / 3.106
\\

Qwen2.5-Omni-3B + UniTalker-B-D3
& 10.449 / 3.222
& 10.632 / 3.156
& 10.544 / 3.153
& 10.494 / 3.419
\\

Qwen2.5-Omni-3B + UniTalker-B-D6
& 9.873 / 3.738
& 10.088 / 3.600
& 9.945 / 3.629
& 10.302 / 3.504
\\

Qwen2.5-Omni-7B + EmoTalk
& 10.554 / 3.273
& 11.215 / 3.628
& 10.513 / 3.381
& 11.180 / 3.641
\\

Qwen2.5-Omni-7B + UniTalker-B-D3
& 10.494 / 3.200
& 10.854 / 3.767
& 10.520 / 3.319
& 10.883 / 3.675
\\

Qwen2.5-Omni-7B + UniTalker-B-D6
& 10.176 / 3.405
& 10.856 / 4.012
& 10.115 / 3.641
& 10.833 / 3.980
\\

\midrule

\textbf{Ex-Omni}
& \textbf{9.233 / 5.385}
& \textbf{9.212 / 5.363}
& \textbf{9.313 / 5.236}
& \textbf{9.239 / 5.403}
\\

\bottomrule
\end{tabularx}

\end{table}

\begin{table}[htbp]
\centering
\scriptsize
\setlength{\tabcolsep}{8pt}

\caption{Controlled downstream comparison with a fixed Ex-Omni response.
All four cascades receive identical Ex-Omni-generated speech; Ex-Omni uses
its jointly trained facial-animation branch. Audio2Face-3D is the pseudo-label
teacher. Best and second-best results are shown in \textbf{bold} and
\underline{underline}, respectively.}
\label{tab:controlled_downstream}

\begin{tabularx}{\textwidth}{
    >{\raggedright\arraybackslash}X
    c c c c
}
\toprule
\multirow{2}{*}{Method}
& \multicolumn{2}{c}{
    Speech-to-Face
    (Sync-D $\downarrow$ / Sync-C $\uparrow$)}
& \multicolumn{2}{c}{
    Text-to-Face
    (Sync-D $\downarrow$ / Sync-C $\uparrow$)}
\\
\cmidrule(lr){2-3}
\cmidrule(lr){4-5}
& AlpacaEval
& CommonEval
& AlpacaEval
& CommonEval
\\
\midrule

Ex-Omni + Audio2Face-3D
& \textbf{8.928 / 5.775}
& \textbf{8.921 / 5.699}
& \textbf{8.913 / 5.669}
& \textbf{8.896 / 5.741}
\\

Ex-Omni + EmoTalk
& 10.758 / 4.321
& 10.416 / 3.710
& 10.534 / 4.319
& 10.418 / 3.490
\\

Ex-Omni + UniTalker-B-D3
& 10.946 / 3.882
& 11.117 / 2.900
& 10.893 / 3.708
& 10.792 / 3.130
\\

Ex-Omni + UniTalker-B-D6
& 10.701 / 4.343
& 10.668 / 3.516
& 10.786 / 4.049
& 10.334 / 3.771
\\

\textbf{Ex-Omni}
& \underline{9.233 / 5.385}
& \underline{9.212 / 5.363}
& \underline{9.313 / 5.236}
& \underline{9.239 / 5.403}
\\

\bottomrule
\end{tabularx}

\end{table}

\subsection{Experimental Results and Analysis}
\paragraph{3D Facial Animation Generation Results.}
\begin{wraptable}{r}{0.48\textwidth}
\centering

\caption{Human preference on CommonEval S2F. Tied first-place votes
are split evenly; tie rates are reported separately.}
\label{tab:human_preference}

\scriptsize
\setlength{\tabcolsep}{5pt}
\renewcommand{\arraystretch}{1.08}

\begin{tabularx}{\linewidth}{
    @{}>{\raggedright\arraybackslash}X c c@{}
}
\toprule
\textbf{Method}
& \textbf{Preference} $\uparrow$
& \textbf{Tie}
\\
\midrule

Ex-Omni + Audio2Face-3D
& \textbf{55.8\%}
& 21.1\%
\\

\textbf{Ex-Omni}
& \underline{41.8\%}
& 21.6\%
\\

Ex-Omni + EmoTalk
& 1.9\%
& 1.8\%
\\

Ex-Omni + UniTalker-B-D3
& 0.3\%
& 1.3\%
\\

Ex-Omni + UniTalker-B-D6
& 0.2\%
& 1.1\%
\\

\bottomrule
\end{tabularx}
\end{wraptable}

Table~\ref{tab:speech2face} first compares complete Ex-Omni outputs with non-teacher cascaded OLLM+S2F systems. Because those cascades use a different upstream speech generator, this table characterizes end-to-end pipeline behavior rather than isolating the facial-animation module. The controlled comparison in Table~\ref{tab:controlled_downstream}, which fixes the upstream Ex-Omni response, is the primary evidence for the native branch. The Audio2Face-3D cascade is expected to rank first because it directly invokes the model that supplies our pseudo-labels. Ex-Omni consistently ranks second and is substantially closer to this teacher cascade than to the other external S2F models. On CommonEval S2F, for example, Ex-Omni differs from the teacher by only 0.291 Sync-D and 0.336 Sync-C; relative to the strongest non-teacher cascade, it improves Sync-D by 1.204 and Sync-C by 1.653. The result indicates that pseudo-label supervision transfers much of the teacher's synchronization behavior into Ex-Omni's native branch, without requiring an external S2F model at inference. The human study in Table~\ref{tab:human_preference} supports the same conclusion: the teacher cascade receives 55.8\% of first-place preference, Ex-Omni receives 41.8\%, and the three non-teacher cascades together receive only 2.4\%. Combined with the face-latency results in Table~\ref{tab:latency}, the main advantage of Ex-Omni is approaching teacher-cascade synchronization while providing a simpler, lower-latency integrated generation pipeline.

\begin{table*}[htbp]
\centering
\scriptsize
\caption{Speech QA comparison on VoiceBench under its standard speech-input setting. $\uparrow$ means higher is better. $*$ denotes results reproduced using open-source code.}
\label{tab:model_rankings}
\setlength{\tabcolsep}{1.8pt}
\begin{tabularx}{\textwidth}{lcccccccccc}
\toprule
Model & AlpacaEval $\uparrow$ & CommonEval $\uparrow$ & WildVoice $\uparrow$ & SD-QA $\uparrow$ & MMSU $\uparrow$ & OBQA $\uparrow$ & BBH $\uparrow$ & IFEval $\uparrow$ & AdvBench $\uparrow$ & Overall $\uparrow$\\
\midrule
Qwen2.5-Omni-7B & \textbf{4.49} & \textbf{3.93} & 2.71$^*$ & \textbf{55.71} & \textbf{61.32} & \textbf{81.10} & \underline{60.80$^*$} & \underline{52.87} & \textbf{99.42} &  \textbf{70.42$^*$} \\
Moshi  & 2.01 & 1.60 & 1.30 & 15.64 & 24.04 & 25.93 & 47.40 & 10.12 & 44.23 & 29.51 \\
VITA-1.0 & 3.38 & 2.15 & 1.87 & 27.94 & 25.70 & 29.01 & 47.70 & 22.82 & 26.73 & 56.48 \\
VITA-1.5 & 4.21 & 3.66 & \underline{3.48} & 38.88 & \underline{52.15} & \underline{71.65} & 55.30 & 38.14 & \underline{97.69} & 64.53 \\
LLaMA-Omni & 3.70 & 3.46 & 2.92 & 39.69 & 25.93 & 27.47 & 49.20 & 14.87 & 11.35 & 41.12 \\
Mini-Omni & 1.95 & 2.02 & 1.61 & 13.92 & 24.69 & 26.59 & 46.30 & 13.58 & 37.12 & 30.42 \\
Mini-Omni2 & 2.32 & 2.18 & 1.79 & 9.31 & 24.27 & 26.59 & 46.40 & 11.56 & 57.50 & 33.49 \\
SLAM-Omni & 1.90 & 1.79 & 1.60 & 4.16 & 26.06 & 25.27 & 48.80 & 13.38 & 94.23 & 35.30 \\
\midrule
\textbf{Ex-Omni} & \underline{4.31} & \underline{3.82} & \textbf{3.49} & \underline{50.71} & 46.03 & 56.70 & \textbf{61.10} & \textbf{55.72} & 87.12 & \underline{65.53} \\

\bottomrule
\end{tabularx}
\end{table*}

\paragraph{Speech QA Results.}
Table~\ref{tab:model_rankings} evaluates whether Ex-Omni retains speech QA capability after introducing native speech--face generation. Ex-Omni achieves the second-best overall score of 65.53, outperforming all compared systems except Qwen2.5-Omni-7B, and ranks first on WildVoice, BBH, and IFEval. This performance is notable because Ex-Omni additionally generates speech-synchronized 3D facial animation, a capability absent from the compared systems. Its remaining gaps on MMSU, OBQA, and AdvBench suggest room for broader knowledge- and safety-oriented instruction tuning, whereas the current 200K-sample S2SF training stage primarily focuses on introducing the new generation modality. Overall, Ex-Omni adds native non-verbal generation while retaining competitive speech understanding and reasoning capabilities.

\begin{wraptable}{r}{0.55\textwidth}
\vspace{-1.0\baselineskip}
\centering
\caption{Speech response quality and speech--text consistency (S-T Consis.) on AlpacaEval and CommonEval. UTMOS is computed on generated speech, while S-T Consis. is measured by WER after transcribing the generated audio with Whisper-Large-V3 and comparing it with the corresponding textual response.}
\label{tab:librispeech_asr}

\scriptsize
\setlength{\tabcolsep}{3pt}
\renewcommand{\arraystretch}{1.05}
\begin{tabularx}{\linewidth}{
    @{}>{\raggedright\arraybackslash}X c c c c@{}
}
\toprule
\multirow{2}{*}{Method}
& \multicolumn{2}{c}{AlpacaEval}
& \multicolumn{2}{c}{CommonEval} \\
\cmidrule(lr){2-3}
\cmidrule(lr){4-5}
& UTMOS $\uparrow$
& S-T Consis. $\downarrow$
& UTMOS $\uparrow$
& S-T Consis. $\downarrow$ \\
\midrule
Qwen2.5-Omni-3B
& 4.174
& \textbf{33.53}
& \underline{4.495}
& \underline{20.10} \\

Qwen2.5-Omni-7B
& \underline{4.502}
& 55.11
& \textbf{4.525}
& 21.72 \\
\midrule

\textbf{Ex-Omni}
& \textbf{4.523}
& \underline{34.22}
& 4.491
& \textbf{3.54} \\
\bottomrule
\end{tabularx}
\vspace{-1.0\baselineskip}
\end{wraptable}

\paragraph{Speech Generation Results.}
Table~\ref{tab:librispeech_asr} evaluates the quality of generated speech and its consistency with the corresponding textual response. Despite jointly learning speech and facial-animation generation, Ex-Omni achieves the best UTMOS on AlpacaEval (4.523) and comparable quality on CommonEval (4.491). Its S-T WER is competitive on AlpacaEval and substantially lower on CommonEval (3.54 versus 20.10--21.72), indicating reliable response-level speech--text alignment. This CommonEval gap is partly affected by response length: Qwen2.5-Omni often produces longer responses, some lasting approximately 90 seconds, while speech beyond roughly 60 seconds is more susceptible to truncation or late-stage errors. Overall, Ex-Omni maintains strong speech quality and speech--text fidelity while additionally generating synchronized facial animation.

\begin{wraptable}{r}{0.62\textwidth}
\vspace{-1.2\baselineskip}
\centering
\caption{Effect of TQGF. QA Score denotes the GPT-based open-ended response score on a 1--5 scale. $\uparrow$/$\downarrow$ indicates that higher/lower is better.}
\label{tab:effect_components}

\scriptsize
\setlength{\tabcolsep}{5.0pt}
\renewcommand{\arraystretch}{1.05}
\begin{tabularx}{\linewidth}{
    @{}>{\raggedright\arraybackslash}l c c c@{}
}
\toprule
Method
& Sync-D$\downarrow$ / Sync-C$\uparrow$
& QA Score$\uparrow$
& UTMOS$\uparrow$ \\
\midrule

\textbf{Ex-Omni}
& \textbf{9.233} / \textbf{5.385}
& \textbf{4.31}
& \textbf{4.523} \\

\hspace{0.2em}\textit{w/o TQGF (only speech tokens)}
& 9.591 / 4.826
& 4.29
& 4.161 \\

\hspace{0.2em}\textit{w/o TQGF (only hidden features)}
& 9.475 / 4.768
& 4.23
& 3.968 \\

\hspace{0.2em}\textit{w/o TQGF (concatenation)}
& 10.209 / 4.043
& 4.27
& 4.136 \\
\bottomrule
\end{tabularx}
\vspace{-1.0\baselineskip}
\end{wraptable}

\paragraph{Effect of TQGF.}
\label{sec:aba_fag}
Table~\ref{tab:effect_components} evaluates the TQGF interface against single-source conditioning and direct feature concatenation. The full model performs best across all metrics. Using only speech tokens reduces Sync-C by 0.559, showing that the temporal scaffold alone lacks sufficient facial context, while using only hidden features reduces Sync-C by 0.617, indicating that contextual representations cannot replace explicit speech-unit timing. Direct concatenation performs worst, increasing Sync-D by 0.976 and reducing Sync-C by 1.342. These results demonstrate that speech tokens and hidden features provide complementary temporal and contextual cues, which TQGF integrates more effectively while preserving response and speech quality.

\paragraph{Ablation Study on Input Modality and Knowledge Retention.}
As shown in Table \ref{tab:modality_retention}, the text-to-text comparison isolates knowledge retention after multimodal training: Ex-Omni decreases by only 0.12 on both AlpacaEval and CommonEval relative to its Qwen3-8B initialization. The additional gap under speech input also includes speech encoding and cross-modal alignment, and therefore should not be interpreted as degradation of the LLM's text reasoning alone. This ablation shows that Ex-Omni retains most of the base model's text capability while acquiring speech understanding and native facial-animation generation.

\begin{table}[htbp]
\centering

\begin{minipage}[t][3.5cm][t]{0.46\textwidth}
\centering

\caption{Input-modality ablation on the open-ended QA subsets.
$\uparrow$ means higher is better.}
\label{tab:modality_retention}
\scriptsize
\setlength{\tabcolsep}{9.0pt}
\renewcommand{\arraystretch}{1.50}
\begin{tabularx}{\linewidth}{lccc}
\toprule
\textbf{Model}
& \textbf{Input}
& \textbf{AlpacaEval} $\uparrow$
& \textbf{CommonEval} $\uparrow$ \\
\midrule

Qwen3-8B
& Text
& \textbf{4.82}
& \textbf{4.51} \\

Ex-Omni
& Text
& \underline{4.70}
& \underline{4.39} \\

Ex-Omni
& Speech
& 4.31
& 3.82 \\
\bottomrule
\end{tabularx}

\end{minipage}
\hfill
\begin{minipage}[t][3.5cm][t]{0.52\textwidth}
\centering

\caption{Latency comparison between Ex-Omni and cascaded
speech-to-face pipelines. $\downarrow$ means lower is better.}
\label{tab:latency}

\scriptsize
\setlength{\tabcolsep}{4.0pt}
\renewcommand{\arraystretch}{1.00}
\begin{tabularx}{\linewidth}{lccc}
\toprule
\textbf{Model}
& \makecell{\textbf{Overall RTF} $\downarrow$}
& \makecell{\textbf{Avg Speech}\\\textbf{TTFT (s)} $\downarrow$}
& \makecell{\textbf{Avg Face}\\\textbf{Latency (s)} $\downarrow$} \\
\midrule

Ex-Omni
& \multirow{4}{*}{2.158}
& \multirow{4}{*}{0.029}
& \textbf{0.012} \\

Ex-Omni + EmoTalk
& & & 0.110 \\

Ex-Omni + UniTalker-B-D3
& & & 0.105 \\

Ex-Omni + UniTalker-B-D6
& & & 0.117 \\
\bottomrule
\end{tabularx}

\end{minipage}

\end{table}

\paragraph{Latency Analysis.}
Table~\ref{tab:latency} reports latency on 100 randomly sampled CommonEval instances using three metrics: Overall RTF, Avg Speech TTFT, and Avg Face Latency. Since all compared systems share the same speech generation backbone, they have identical Overall RTF (2.158) and Avg Speech TTFT (0.029s). The main difference lies in face generation latency: Ex-Omni requires only 0.012s, while cascaded pipelines with task-specific S2F models require 0.105--0.117s. This indicates that the native facial-animation branch has lower face-generation latency than invoking a separate downstream S2F model. The overall RTF is still above real time under the tested NVIDIA H20 GPU, suggesting that the main bottleneck remains the Qwen3-8B semantic reasoning backbone rather than the facial animation branch.

\section{Conclusion}
In this paper, we introduced Ex-Omni, a framework that extends OLLMs with native speech-accompanied 3D facial animation generation. To address the mismatch between token-level semantic reasoning and fine-grained facial motion, Ex-Omni decouples high-level understanding from modality-specific temporal synthesis through discrete speech-unit scaffolding and a task-specific token-as-query gated fusion interface. We further constructed InstructS2SF-1200K to provide large-scale weak speech--face supervision for both synthesis and dialogue-oriented scenarios. Experiments show that Ex-Omni preserves substantial speech understanding and generation capability, improves over the evaluated non-teacher cascades in controlled downstream comparisons, and approaches the Audio2Face-3D teacher cascade in both SyncNet and human preference while avoiding an external S2F model at inference.

\section*{Limitations}
Ex-Omni has four main limitations. First, long speech responses produce lengthy frame-level facial sequences, increasing computational and memory costs and challenging long-range consistency. Second, the overall real-time factor remains above 1.0, and the latency advantage is limited to the facial-animation branch rather than the complete pipeline. Third, the compact ARKit-52 representation may underrepresent subtle affective and person-specific facial motions~\cite{DBLP:journals/corr/abs-2508-16401}. Finally, reliance on Audio2Face-3D pseudo-labels may introduce teacher-specific biases.

\bibliographystyle{assets/plainnat}
\bibliography{citation}

\clearpage
\appendix

\section{Implementation Details}
\paragraph{Data Construction.}
\label{appendix: data_construction}
In Stage I, about 720K ASR samples are primarily sampled from the Emilia \citep{emilialarge} corpus. We further incorporate the train-clean-100, train-clean-360, and train-other-500 subsets of LibriSpeech \cite{DBLP:conf/icassp/PanayotovCPK15}, together with WenetSpeech \cite{DBLP:conf/icassp/ZhangLGSYXXBCZW22} training data whose confidence scores exceed 0.95.

In Stage II, we construct 1M TTSF samples. Specifically, text prompts sampled from Emilia \cite{emilia, emilialarge} are converted into speech using Qwen3-TTS \cite{hu2026qwen3} with a single, unified speaker identity. High-quality motion-capture facial animation data are scarce, and existing public datasets typically contain only a few thousand samples with limited coverage of speech content. We therefore adopt NVIDIA Audio2Face-3D \cite{DBLP:journals/corr/abs-2508-16401}, an open-weights model trained on large-scale professionally captured facial motion, as a teacher to generate blendshape pseudo-labels from the synthesized speech. Concretely, we first run the off-the-shelf NVIDIA Audio2Emotion estimator \cite{DBLP:journals/corr/abs-2508-16401} on the synthesized audio to obtain a 5D compound emotion vector over anger, disgust, sadness, joy, and fear, where an all-zero vector is treated as neutral. Audio2Emotion is used only for dataset construction and is not trained by us. This emotion vector is then used as the conditional input to Audio2Face-3D to generate the corresponding blendshape sequence. These teacher-generated blendshape sequences are used as weak supervision signals for speech-blendshape co-pretraining, rather than being treated as deterministic facial ground truth. 

In Stage III, we construct 200K dialogue-based S2SF samples based on InstructS2S-200K \cite{DBLP:conf/iclr/FangGZMZ025}. To improve speech quality and consistency, we reconstruct the target speech using Qwen3-TTS, and then use Audio2Face-3D \cite{DBLP:journals/corr/abs-2508-16401} to generate corresponding blendshape pseudo-labels. As in Stage II, we first estimate a 5D compound emotion vector from the audio with Audio2Emotion and use it as the conditional input to Audio2Face-3D. In practice, we observe that most speech responses in QA-style dialogue are close to neutral under this pipeline, so the resulting pseudo-labels mainly emphasize accurate mouth articulation and lip-speech synchronization rather than exaggerated facial expressions. As in Stage II, these teacher-generated blendshape sequences serve as weak supervision signals, which is a more appropriate formulation given the one-to-many nature of speech-to-blendshape mapping. To preserve capabilities acquired in earlier stages, we additionally include 10K ASR replay data and 10K TTSF replay data during this stage.
Standalone audio-to-animation systems such as xADA~\cite{taylor2025xada} address a different optimization setting from Ex-Omni, which jointly learns text generation, speech-unit generation, and facial regression from heterogeneous supervision. In reduced-data pilot experiments, sparsifying the animation-supervised subset caused larger loss fluctuations and unstable joint optimization. We therefore retain dense pseudo-label coverage during speech--blendshape co-pretraining; this choice concerns supervision scale rather than treating the pseudo-labels as unique ground truth.

\paragraph{Hyperparameter.}
All experiments are conducted on a machine equipped with 8 NVIDIA H20 GPUs, each with 96 GB of memory. We use CUDA~12.6, PyTorch~2.7.0 and Python 3.10 for model training and evaluation. The detailed hyperparameters of the three-stage training schedule are shown in Table~\ref{tab:hyperparameters}.

\begin{table}[htbp]
\centering
\scriptsize
\setlength{\tabcolsep}{5.0pt}
\caption{The detailed training setup for Ex-Omni across the three training stages.}
\label{tab:hyperparameters}
\begin{tabularx}{0.48\textwidth}{>{\raggedright\arraybackslash}l c c c}
\toprule
\textbf{Hyperparameter} & \textbf{I} & \textbf{II} & \textbf{III} \\
\midrule
epoch & 1 & 2 & 2 \\
effective batch size & 128 & 128 & 32 \\
optimizer & AdamW & AdamW & AdamW \\
warmup ratio & 0.3 & 0.1 & 0.1 \\
Gradient Accumulation & 1 & 2 & 4 \\
lr of Speech Encoder & 0 & 0 & 0 \\
lr of Speech Projector & $1 \times 10^{-3}$ & 0 & $2 \times 10^{-5}$ \\
lr of LLM & 0 & 0 & $2 \times 10^{-6}$ \\
\shortstack[l]{lr of Blendshape-Aware\\Speech Unit Generator} & 0 & $1 \times 10^{-4}$ & $2 \times 10^{-5}$ \\
lr of blendshape decoder & 0 & $1 \times 10^{-4}$ & $2 \times 10^{-5}$ \\
lr of Speech Decoder & 0 & 0 & 0 \\
freeze Speech Encoder & \checkmark & \checkmark & \checkmark \\
freeze Speech Projector & \ding{55} & \checkmark & \ding{55} \\
freeze LLM & \checkmark & \checkmark & \ding{55} \\
\shortstack[l]{freeze Blendshape-Aware\\Speech Unit Generator} & \checkmark & \ding{55} & \ding{55} \\
freeze blendshape decoder & \checkmark & \ding{55} & \ding{55} \\
freeze Speech Decoder & \checkmark & \checkmark & \checkmark \\
\bottomrule
\end{tabularx}
\end{table}

\paragraph{Component Provenance.}
The speech encoder is Whisper-Large-V3 \citep{DBLP:conf/icml/RadfordKXBMS23} and remains frozen throughout training. A two-layer MLP speech projector maps its downsampled features into the Qwen3-8B \citep{DBLP:journals/corr/abs-2505-09388} embedding space. The speech-unit generator with blendshape co-supervision uses a Qwen3-0.6B \citep{DBLP:journals/corr/abs-2505-09388} backbone, while discrete unit tokenization and waveform decoding follow GLM-4-Voice \citep{DBLP:journals/corr/abs-2412-02612}. The GLM-4-Voice waveform decoder remains frozen, and no waveform-level decoder is trained by Ex-Omni. As clarified in the main text, $\mathrm{Emb}_{\theta_L}$ and $\mathrm{Emb}_{\theta_U}$ are distinct input embedding layers associated with the Qwen3-8B LLM and the GLM-4-Voice-derived unit-generation pathway, respectively. 

\paragraph{Convergence and Early Stopping.}
Stage II contains 1M samples totaling 2814.57 hours, so two epochs already correspond to large-scale co-pretraining. Its training loss decreases smoothly and reaches a plateau. Stage III uses a much smaller LLM learning rate together with ASR and TTSF replay. Its mixed-objective loss fluctuates because different batches expose different supervision masks, but the fluctuation amplitude decreases and the loss stabilizes. We select two epochs by early stopping because continuing joint adaptation can degrade knowledge retained from the base LLM, consistent with the text-only QA trade-off reported in Table~\ref{tab:modality_retention}.

\section{Additional Experimental Results}
\paragraph{Temporal Motion Diagnostics.}
To supplement SyncNet and human preference, Table~\ref{tab:motion_dynamics} reports first- and second-order temporal differences of the generated blendshape sequences. For blendshape vector $y_t$, velocity and acceleration are defined as $v_t=y_t-y_{t-1}$ and $a_t=v_t-v_{t-1}$. The \textsc{Abs} columns average coefficient-wise absolute magnitudes, and the $L_2$ columns average vector norms. These values describe motion scale and temporal variation; a very small value may indicate either smooth motion or under-animation. We use them as diagnostics alongside the human preference study.

\begin{table}[htbp]
\centering
\scriptsize
\caption{Velocity and acceleration diagnostics for generated ARKit-52 blendshape sequences. These statistics characterize motion dynamics and should be interpreted together with SyncNet, human preference, and qualitative results.}
\label{tab:motion_dynamics}
\setlength{\tabcolsep}{13pt}
\begin{tabularx}{\textwidth}{>{\raggedright\arraybackslash}X c c c c}
\toprule
Method & Vel-Abs & Vel-$L_2$ & Acc-Abs & Acc-$L_2$ \\
\midrule
Ex-Omni + UniTalker-B-D3 & 0.006834 & 0.074958 & 0.005849 & 0.061824 \\
Ex-Omni + UniTalker-B-D6 & 0.006403 & 0.078124 & 0.005402 & 0.062921 \\
Ex-Omni + EmoTalk & 0.012511 & 0.141280 & 0.010173 & 0.110909 \\
Ex-Omni + Audio2Face-3D & 0.005135 & 0.080290 & 0.002846 & 0.043408 \\
\textbf{Ex-Omni} & 0.009687 & 0.138971 & 0.009532 & 0.131706 \\
\bottomrule
\end{tabularx}
\end{table}

\paragraph{Case Study of 3D Facial Animation Generation.}
We provide qualitative examples in Appendix Figure~\ref{fig:case_study}, where three representative cases are rendered with two different templates: the NVIDIA-provided Claire template and a commercially purchased template with realistic skin and hair materials. We compare Ex-Omni with three cascaded variants, namely Ex-Omni+EmoTalk, Ex-Omni+UniTalker-B-D3, and Ex-Omni+UniTalker-B-D6. To isolate the effect of the facial animation module, all cascaded variants use the same text and speech responses generated by Ex-Omni as input, ensuring that the linguistic content, speech duration, and acoustic realization are controlled across methods. Therefore, the visual differences mainly reflect the quality of the predicted facial animation rather than differences in upstream response generation. As shown in the figure, Ex-Omni captures finer articulatory details more consistently than the cascaded alternatives. In Case 1, when pronouncing ``quote'', the mouth generated by Ex-Omni moves toward a rounded shape, while the other methods fail to capture this phoneme-related transition. In Case 2, after the speech ends, Ex-Omni closes the mouth naturally, whereas the cascaded methods tend to leave the mouth partially open. In Case 3, both Ex-Omni and Ex-Omni+UniTalker-B-D6 produce a large mouth opening for the pronunciation of ``hard'', while the other cascaded variants miss this articulation cue. These differences are more evident in the supplementary videos. Overall, Ex-Omni produces more detailed and stable mouth dynamics while maintaining lip-speech alignment, suggesting that jointly modeling speech units and blendshape generation helps preserve fine-grained temporal cues that may be weakened when facial animation is generated by a separate downstream model.

\begin{figure}[!t]
\centering
\includegraphics[width=0.60\linewidth]{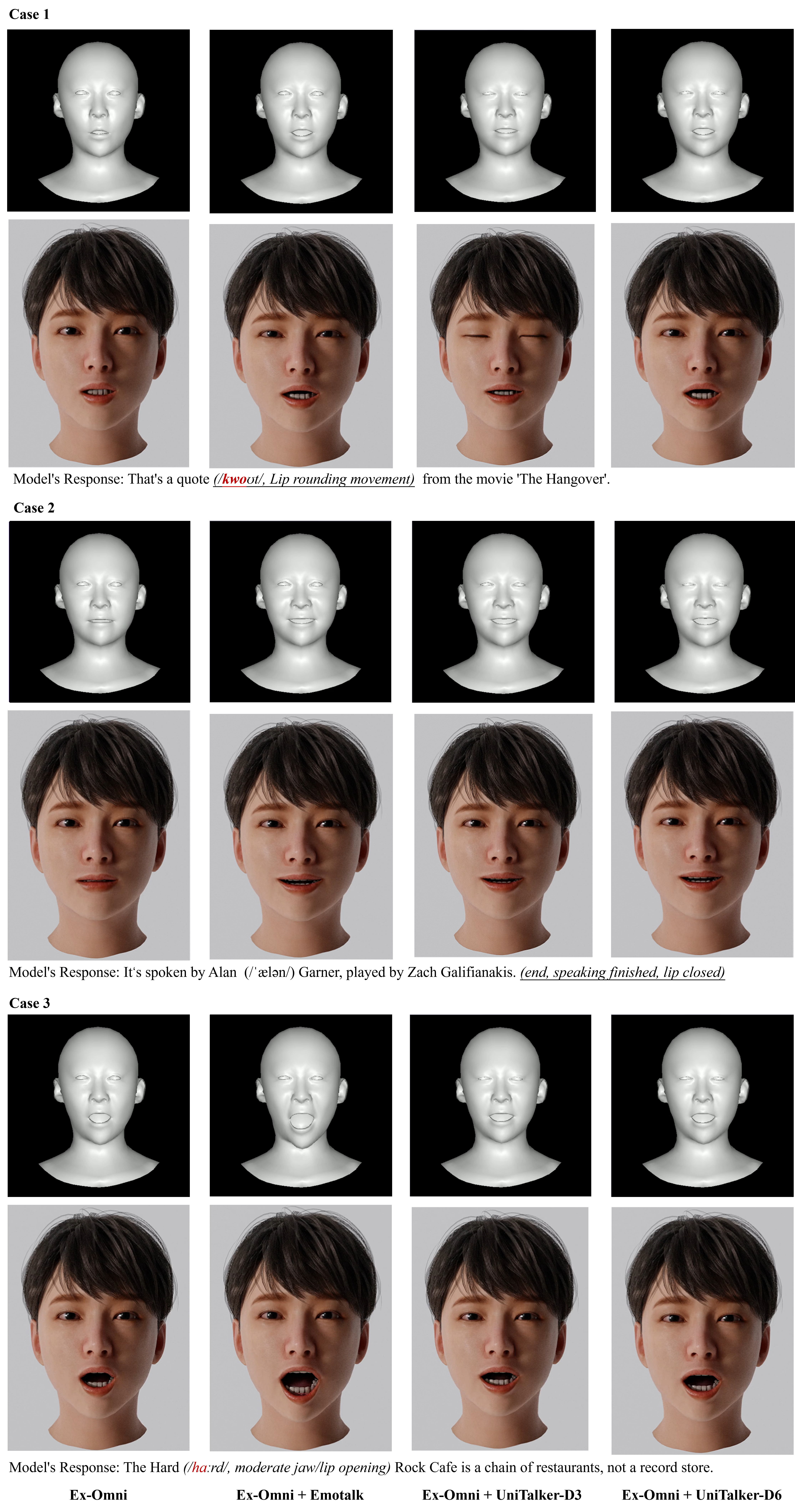}
\caption{Case study on 3D facial animation generation. Three representative cases are rendered using both the NVIDIA-provided Claire template and a commercially purchased template with realistic skin and hair materials. All cascaded baselines use the same Ex-Omni-generated text and speech responses as input.}
\label{fig:case_study}
\end{figure}

\end{document}